%% file: iclr2027_conference.tex
\documentclass{article} 
\usepackage{iclr2027_conference,times}

\input{math_commands.tex}

\usepackage{hyperref}
\usepackage{url}
\usepackage{booktabs}
\usepackage{graphicx}
\usepackage[table]{xcolor}

\title{Precise Editing and Flexible Referencing \\ for Interactable Worlds}

\author{
    \textbf{Xinyao Liao}$^{1,2}$\quad
    \textbf{Xianfang Zeng}$^{2}$\footnotemark[1]\quad
    \textbf{Zhu Liang}$^{2}$\quad
    \textbf{Zhoujie Fu}$^{1}$\quad
    \textbf{Qianxun Xu}$^{2}$\quad
    \textbf{Jiachi Liu}$^{1}$\quad\\[1ex]
    \textbf{Gang Yu}$^{2}$\footnotemark[2]\quad
    \textbf{Guosheng Lin}$^{1}$\footnotemark[2]\\[1ex]
    $^{1}$Nanyang Technological University \quad
    $^{2}$StepFun\\[1ex]
}

\iclrfinalcopy 
\begin{document}
\maketitle
\renewcommand{\thefootnote}{\fnsymbol{footnote}}
\footnotetext[1]{Xianfang Zeng is the project leader.}
\footnotetext[2]{Corresponding authors: skicy@outlook.com, gslin@ntu.edu.sg}
\renewcommand{\thefootnote}{\arabic{footnote}}
\input{section/0_abstract}    
\input{section/1_introduction}
\input{section/2_related_work}
\input{section/3_methodology}
\input{section/4_experiments}
\input{section/5_conclusion}
\bibliography{iclr2027_conference}
\bibliographystyle{iclr2027_conference}
\end{document}

%% file: math_commands.tex
\usepackage{amsmath,amsfonts,bm}

\def\eqref#1{equation~\ref{#1}}

\def\1{\bm{1}}

\DeclareMathAlphabet{\mathsfit}{\encodingdefault}{\sfdefault}{m}{sl}
\SetMathAlphabet{\mathsfit}{bold}{\encodingdefault}{\sfdefault}{bx}{n}



%% file: section/0_abstract.tex
\begin{abstract}
We present EditWorld, a video world model for precise editing and flexible referencing in interactable worlds. Existing video world models primarily focus on navigation, letting users explore generated worlds but offering limited control over how existing world content is modified. EditWorld extends world modeling from exploration to precise modification by streaming editing instructions and reference images during autoregressive generation.
To support these capabilities, EditWorld introduces Gated Causal Attention for temporally varying editing conditions and reference images, together with a Sparse Context mechanism that maintains a bounded historical context for long-horizon inference. We further adopt joint autoregressive and bidirectional training with annealed self-resampling, and construct a dedicated data synthesis and annotation pipeline that provides supervision for world editing. We also present WBench-Editing to systematically evaluate streaming world editing capabilities.
EditWorld achieves the best overall performance on WBench-Editing with an overall score of 73.8 and an editing score of 80.0, substantially outperforming existing methods on editing-related metrics.
\url{https://github.com/leoisufa/EditWorld}
\end{abstract}

%% file: section/1_introduction.tex
\section{Introduction}
Video world models, driven by recent advances in autoregressive video generation, have emerged as a promising substrate for world exploration \citep{team2026advancing,gao2026infinite,sun2025worldplay}, game generation \citep{li2025hunyuan,tang2025hunyuan}, and embodied simulation \citep{team2606kairos}. These models autoregressively generate videos in response to streaming user inputs, such as actions and prompts. Existing video world models, however, have primarily emphasized navigation, focusing on faithful control of camera trajectories and user actions. More recent efforts have begun to extend this capability toward text-driven event generation. YUME 1.5 \citep{mao2026yume1} supports text-controlled world events, HY-WorldPlay 1.5 \citep{sun2025worldplay} enables promptable events across diverse scenes, and LingBot-World and LingBot-World 2.0 \citep{team2026advancing,gao2026infinite} further expand the range of text-driven events and interactive actions. DreamX-World \citep{team2026dreamx} additionally introduces composable event control through event instruction tuning. Despite this progress, existing approaches mainly focus on triggering or generating new events, rather than precisely modifying specified content already present in the world, such as addition, removal, replacement, and stylization. Moreover, flexible incorporation of content from reference images remains insufficiently explored. XGEN-JING \citep{xgen_jing} and ABot-World \citep{jiang2026abot} support identity conditioning from initial references, but do not enable users to interactively and flexibly inject content from different reference images into the generated world over time. As a result, although existing world models increasingly support navigation, text-driven events, and reference-based conditioning, they still lack precise control over editing existing world content and flexible integration of reference images throughout interaction.

To this end, we present \textbf{EditWorld}, a video world model for precise editing and flexible referencing in interactable worlds. EditWorld enables users to continuously modify world content through streaming editing instructions and to flexibly incorporate information from reference images during generation. Specifically, our model is built upon an autoregressive video generation framework in which images or videos are used as the world prior, while camera poses, textual prompts, and reference images are incorporated as conditioning signals for navigation and modification. Starting from LingBot-World-Base \citep{team2026advancing}, a bidirectional video world model, Gated Causal Attention is introduced to support streaming editing instructions and reference images while preserving causal video generation, and the model is further adapted to autoregressive generation through teacher-forcing \citep{williams1989learning} training. Joint autoregressive and bidirectional objectives \citep{gao2026infinite} are adopted to improve condition-following capability. To prevent the video context from growing unboundedly with video length, a Sparse Context mechanism is designed to constrain the historical context to a fixed budget during both training and inference. Furthermore, self-resampling \citep{guo2025end} is adopted to mitigate error accumulation during autoregressive rollouts and improve fidelity and stability over long-horizon generation. On the data side, a dedicated pipeline is developed. Based on public navigation and editing datasets \citep{li2026sekai,wang2026spatialvid,zhou2025omniworld,he2025openve,bai2026scaling}, multiple off-the-shelf methods are leveraged to construct editable world data with detailed editing annotations and reference images, providing supervision for fine-grained content modification and reference-guided generation.

Since existing world model benchmarks primarily evaluate navigation capabilities \citep{wu2026omni,lu2026current,ding2026playworld,xu2026worldmark}, we introduce WBench-Editing, a sub-benchmark of WBench \citep{ying2026wbench} designed to systematically evaluate and compare the streaming editing capabilities of existing world models. WBench-Editing consists of approximately 150 cases, each spanning 240--480 frames and involving one to three streaming editing instructions, with a subset additionally incorporating reference images. Our model achieves the best overall performance on WBench-Editing, with an overall score of 73.8 and an editing score of 80.0, substantially outperforming existing methods in editing capability. To further demonstrate that our approach preserves strong general world modeling capabilities, we also report results on the original WBench, where our model achieves performance comparable to several commercial world models.

In summary, this paper makes the following contributions:
\begin{itemize}
    \item A video world model for \textbf{precise editing and flexible referencing} is developed, enabling users to continuously modify world content through streaming edit instructions and incorporate content from reference images.
    \item WBench-Editing, a sub-benchmark of WBench, is introduced to systematically evaluate streaming world editing capabilities.
    \item Our model achieves the best overall performance on WBench-Editing, with a substantial advantage in editing capability, while maintaining competitive performance on WBench.
\end{itemize}

%% file: section/2_related_work.tex
\section{Related Work}
\noindent\textbf{Video World Models.}
Video world models aim to create infinite worlds with versatile interactions. YUME \citep{mao2025yume}, ASTRA \citep{zhu2026astra}, and Matrix-Game \citep{zhang2025matrix} pioneer interactive video world modeling by generating videos from input images and enabling world exploration through action or camera-trajectory control.
Subsequent efforts have focused on improving viewpoint-control accuracy. HY-World 1.5 \citep{sun2025worldplay} utilizes PRoPE, which explicitly incorporates camera poses as positional priors for video tokens. LingBot-World \citep{team2026advancing} encodes camera poses as Plücker features to inject token-wise spatial information. The Hunyuan-GameCraft series \citep{li2025hunyuan, tang2025hunyuan} maps keyboard and mouse inputs into a shared camera representation space, while the Matrix-Game series \citep{zhang2025matrix, he2025matrix, wang2026matrix} enables frame-level keyboard and mouse conditioning. DreamX-World \citep{team2026dreamx} introduces E-PRoPE with relative frustum-based encoding, and Wonder \citep{xu2026wonder} further improves camera-pose control through a dense coordinate field.
Some works improve long-horizon world consistency by introducing specialized memory modules to mitigate appearance drift when revisiting previously explored locations. AlayaWorld \citep{team2026alayaworld} combines an explicit 3D reprojection cache with compressed representations of recent frames. WorldKV \citep{yi2026worldkv} retrieves evicted KV-cache chunks according to the current camera viewpoint, while StableWorld \citep{yang2026stableworld} employs Dynamic Frame Eviction to discard frames that have accumulated visual drift.
Generation efficiency is another key factor in the practicality of video world models. SANA-WM \citep{zhu2026sana} leverages linear attention for real-time generation over minute-long horizons, while minWM \citep{zhao2026minwm} and SolarWM \citep{huang2026solarwm} provide a fully open-source end-to-end framework for efficient world modeling. ABot-World-0 \citep{jiang2026abot} improves generation speed through a lightweight VAE decoder and efficient attention mechanisms, while MoWorld \citep{moxin2026moworld} demonstrates real-time interactive world modeling on NPUs.
Beyond navigation and action control, recent works have also explored text-conditioned event generation and initial reference-based identity conditioning. YUME 1.5 \citep{mao2026yume1} improves the accuracy of text-controlled event generation, while LingBot-World 2.0 \citep{gao2026infinite} supports multiple text-driven events. XGEN-JING \citep{xgen_jing} and ABot-World-0 \citep{jiang2026abot} further introduce reference-based identity conditioning, enabling identity information to be preserved during world generation.

\noindent\textbf{Autoregressive Video Generation.}
Autoregressive video generation serves as a key technology for interactable video world models. Diffusion Forcing \citep{chen2024diffusion} assigns an independent noise level to each frame, unifying next-frame prediction with full-sequence diffusion. Self-Forcing \citep{huang2026self} addresses exposure bias by performing autoregressive rollouts with KV caching during training, such that each frame is conditioned on previously generated outputs. Building on this paradigm, Self-Forcing++ \citep{cui2026self} samples training clips from self-generated long videos and leverages knowledge from a teacher model, while Self Gradient Forcing \citep{zhuang2026self} enables gradients from future predictions to propagate through historical KV states. Context Forcing \citep{chen2026context} further replaces short-context teachers with long-context supervision. Self-Resampling \citep{guo2025end} performs end-to-end training from scratch while explicitly simulating inference-time errors during training, whereas Reward-Forcing \citep{zhang2026reward} replaces teacher supervision with reward signals. Causal Forcing \citep{zhu2026causal} identifies a theoretical inconsistency in distilling autoregressive students from bidirectional teachers, arising from violations of frame-level injectivity. Causal Forcing++ \citep{zhao2026causal} extends this framework to one/two-step sampling per frame and identifies initialization as a critical bottleneck.

%% file: section/3_methodology.tex
\begin{figure}[t]
\centering
\includegraphics[width=0.95\linewidth]{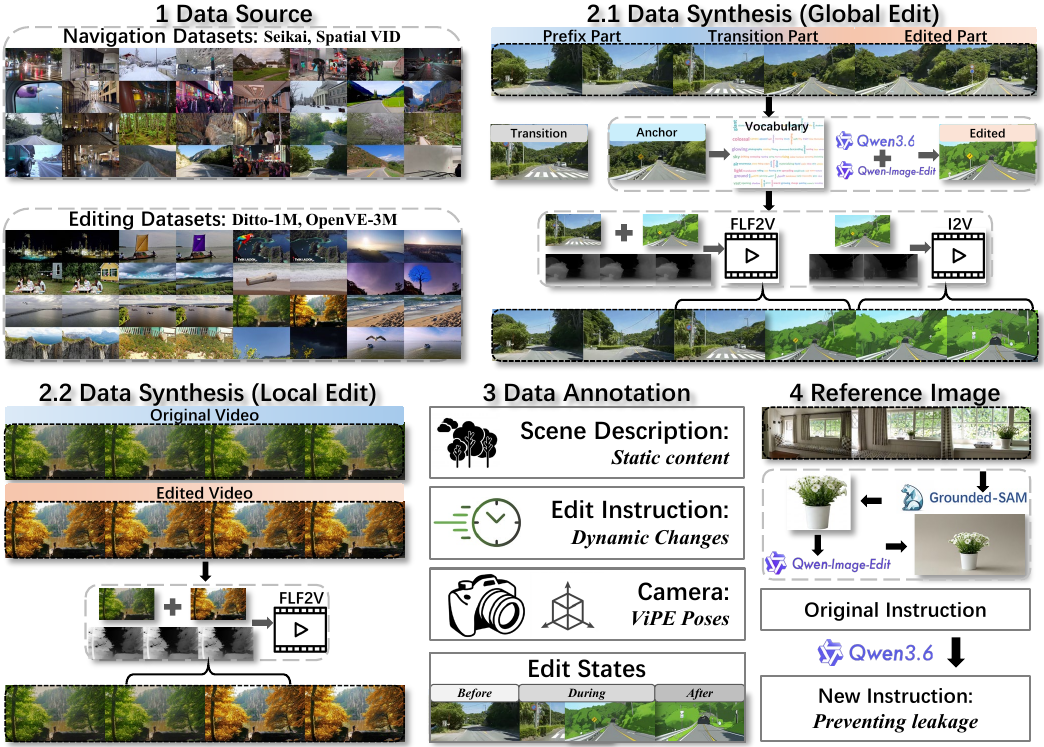}
\caption{Data pipeline overview. (1) Our raw data is collected from two main sources: open-source navigation datasets and video editing datasets. (2) Sections 2.1 and 2.2 illustrate the synthesis pipelines for constructing global and local editable world data from the source videos. (3) Each synthesized training sample is annotated with four components: scene description, editing instruction, editing state, and camera poses. (4) Reference images are further extracted from the synthesized videos, together with rewritten editing instructions to prevent semantic leakage.}
\label{fig:datapipeline}
\end{figure}

\section{Methodology}
\subsection{Data Pipeline}
Existing world model datasets typically consist of navigation videos collected in large-scale environments, providing limited supervision for modifying world content. In contrast, public video editing datasets usually contain only source--edited video pairs and do not explicitly model the smooth temporal transition from the original state to the edited state. To train a video world model for precise editing and flexible referencing in interactable worlds, we develop a data pipeline for synthesizing long-horizon navigation videos with temporally grounded editing events and reference-image conditioning. As shown in Figure \ref{fig:datapipeline}, the source data are drawn from two categories: Sekai \citep{li2026sekai}, OmniWorld \citep{zhou2025omniworld}, and SpatialVID \citep{wang2026spatialvid} are used as navigation data, while Ditto-1M \citep{bai2026scaling} and OpenVE-3M \citep{he2025openve} are used as editing data. We broadly categorize editing operations into global editing and local editing. Global editing refers to holistic changes in the appearance of the world, such as changes in weather, season, time, illumination, color tone, and artistic style, whereas local editing refers to localized modifications to individual elements, including addition, removal, and modification. Separate synthesis pipelines are designed for these two editing categories. Each synthesized video is further annotated with scene descriptions, editing instructions, editing states, camera trajectories, and optional reference images.

\noindent\textbf{Global Editing Data.}
The goal of global editing data is to provide supervision for holistic world transitions. Given a raw navigation video, a global editing instruction is first constructed, and an anchor frame is selected, after which the edit is propagated across the video. We build an editing vocabulary containing approximately 200 primary global editing attributes and 100 auxiliary attributes. For each video, Qwen3.6-27B \citep{qwen3.6-27b} is used to sample one primary attribute and one compatible auxiliary attribute according to the video content. The selected attribute combination is then instantiated into a concrete global editing prompt that is consistent with the current scene. The anchor frame and the editing instruction are subsequently provided to Qwen-Image-Edit \citep{wu2025qwen} to generate the edited anchor frame.

Given a high-quality edited anchor frame, the single-frame edit is extended to the full video through a smooth temporal transition. Let the anchor-frame position in the original video be denoted by $F_a$. We determine the starting point of the transition segment, denoted by $F_t$, and preserve the original video before $F_t$ as the unedited prefix. For the transition segment, the original frame at $F_t$ is used as the first-frame condition, while the edited anchor frame at $F_a$ is used as the last-frame condition. A VLM is then prompted to describe the smooth state transition between the two boundary frames. The transition prompt and the two boundary frames are fed into a depth-controlled first-last-frame-to-video model, Wan2.2-FLF2V-A14B-Control \citep{aigc_apps_VideoX_Fun_2026}, to synthesize the transition segment. To preserve the spatial structure and camera motion of the original video, depth maps from the corresponding temporal interval are extracted and used as structural control signals, reducing undesired drift in camera motion and scene geometry. For the segment after $F_a$, we employ the depth-controlled image-to-video model Wan2.2-I2V-A14B-Control \citep{aigc_apps_VideoX_Fun_2026}, using the edited anchor frame as the initial-frame condition and the depth sequence extracted from the original video after $F_a$ as the control signal. Finally, the unedited prefix, generated transition segment, and edited continuation are concatenated to form the complete video.

\noindent\textbf{Local Editing Data.}
Local editing data is designed to teach the model element-level modification capabilities, including adding, removing, or replacing specific elements in the world. Beyond object-level operations, local editing also covers localized attribute changes, such as modifications to color, material, shape, and state. This portion of the dataset is primarily constructed from publicly available video editing datasets that provide source videos, edited videos, and corresponding editing instructions. To improve data quality, we first apply a two-stage VLM-based filtering pipeline to the collected video pairs. In the first stage, the VLM determines whether the difference between the source and edited videos corresponds to a local editing operation. In the second stage, semantic consistency is evaluated together with the overall video quality. For each video pair that passes filtering, a source frame $F_s$ is selected from the original video and a target frame $F_t$ from the edited video. The source frame is required to clearly present the target element before editing, while the target frame should fully capture the desired post-edit state. The temporal interval between these two endpoint frames is treated as the transition segment. We then provide $F_s$, $F_t$, and the original editing instruction to a VLM to generate a transition prompt describing the required visual change. The transition segment is synthesized using the same depth-controlled video generation model as in the global editing pipeline. Finally, the source-video segment before $F_s$, the generated transition segment, and the edited-video segment after $F_t$ are concatenated to form the final video.

\noindent\textbf{Data Annotation.}
Our annotations consist of four components: scene description, editing instruction, editing state, and camera poses. The scene description captures the static and invariant content of the scene. To generate this annotation, the synthesized video together with its editing instruction is provided to a VLM, which is prompted to describe the scene while excluding elements affected by the editing operation. Although editing instructions are already obtained during the preceding synthesis process, they may be inaccurate or incomplete. We therefore provide the original editing instruction, the synthesized video, and the generated scene description to the VLM, and prompt it to produce a more accurate and detailed editing instruction while avoiding redundancy or conflicts with the scene description. This process decouples the textual conditioning of each video into two complementary components: the scene description, which represents static content, and the editing instruction, which specifies dynamic changes. For editing-state annotations, each video is divided into three temporal states: \textit{before}, \textit{during}, and \textit{after}. This design provides fine-grained temporal supervision for chunk-level editing control. The video is partitioned into chunks, and a VLM is used to assign an editing state to each chunk. Finally, ViPE \citep{huang2025vipe} is used to re-estimate the camera intrinsics and extrinsics for all training samples, providing consistent camera annotations.

\noindent\textbf{Reference Images.}
A subset of the training data is further augmented with reference images to teach the model how to incorporate content from external references into the generated world. For local editing data, Grounded-SAM \citep{ren2024grounded} is used to segment the target object from the selected reference frame. The resulting segmentation is then provided to Qwen-Image-Edit \citep{wu2025qwen} to repair incomplete or imperfect regions and produce the final reference image. For global editing data, the previously generated edited anchor frame is used as the basis for reference construction. The anchor frame is provided to the image editing model, while a VLM generates an instruction that alters the scene, layout, and environment while preserving the target editing attributes, such as weather, style, or time. This process produces a reference image that retains the desired editing attributes while differing from the original world in scene-level content. We further rewrite the corresponding editing instructions for reference-conditioned samples. Specifically, descriptions of content already conveyed by the reference image are removed from the textual instruction to prevent semantic leakage. As a result, the model cannot rely solely on text to recover the target content and is instead encouraged to extract and incorporate the relevant information from the reference image.

\subsection{EditWorld}
\begin{figure}[t]
\centering
\includegraphics[width=0.95\linewidth]{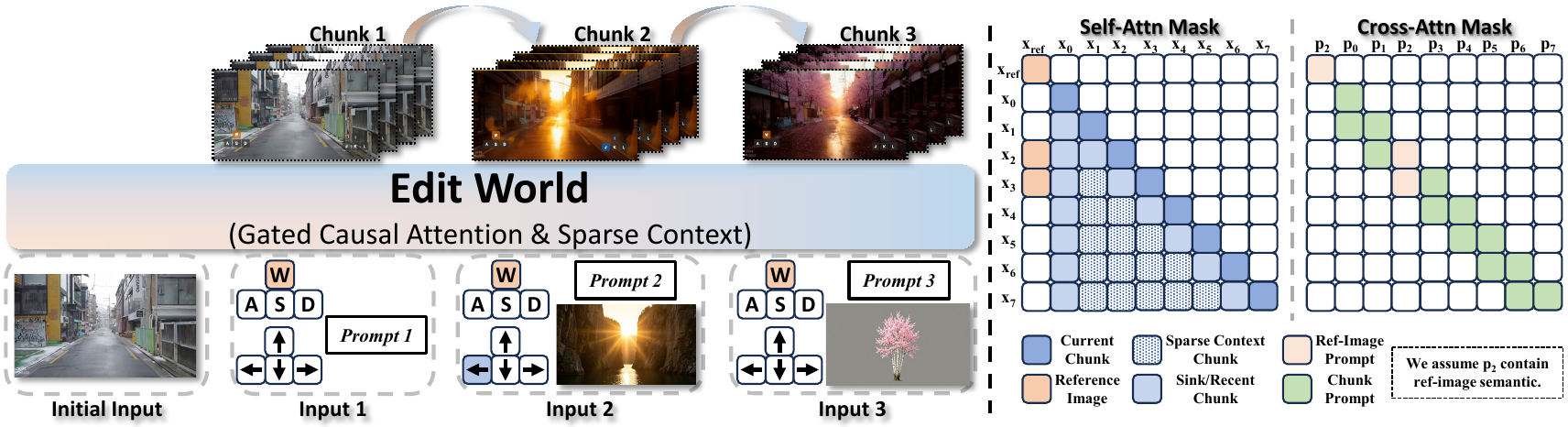}
\caption{EditWorld. The left panel illustrates how our model uses an initial image as the world prior and autoregressively generates and edits the world in response to user-provided actions, textual prompts, and reference images. The right panel visualizes the attention patterns of our Gated Causal Attention and Sparse Context mechanisms. For simplicity, we illustrate the case with one sink chunk, one recent chunk, one preceding prompt, and one reference image.}
\label{fig:model}
\end{figure}

As shown in the left part of Figure \ref{fig:model}, our world model takes an initial frame as the world prior and autoregressively generates an interactable world in response to a stream of user inputs, including textual prompts, actions, and reference images. To support precise editing and flexible referencing during interaction, world generation is formulated as a causal video generation process, where each video chunk is conditioned on preceding observations and the user inputs available up to the current time step. Let $\mathcal{V} = \{x_1, x_2, \ldots, x_T\}$ denote a sequence of video chunks, where $x_t \in \mathbb{R}^{L \times H \times W \times C}$ represents a chunk of $L$ frames at time index $t$, and let $\mathcal{C} = \{c_1, c_2, \ldots, c_T\}$ denote the corresponding sequence of user inputs. Under the causal formulation, the generation process is factorized as
\begin{equation}
    p_{\theta}(x_{1:T} \mid c_{1:T})
    =
    \prod_{t=1}^{T} p_{\theta}(x_t \mid x_{<t}, c_{\leq t}).
\end{equation}
Here, $\theta$ denotes the model parameters. Causality is enforced through both the model architecture and the training curriculum, as detailed in the following sections.

\subsubsection{Causal Video Model}
We train a causal video generation model for multi-condition controllable world generation. Our model is built upon LingBot-World-Base \citep{team2026advancing}, a bidirectional video world model. As illustrated in the right part of Figure \ref{fig:model}, Gated Causal Attention is introduced to support streaming editing instructions and reference images during autoregressive generation while preserving temporal causality and continuity. In parallel, we design a Sparse Context mechanism that constrains the video latent context to a fixed budget during inference.

\noindent\textbf{Gated Causal Attention.}
As described in the data pipeline section, each video is annotated with a scene description, an editing instruction, and chunk-wise editing states. These annotations are used to construct the gated cross-attention mechanism. For each video chunk $x_t$, the corresponding prompt tokens $p_t$ always contain the scene description prompt $p_{\text{scene}}$, while the editing instruction prompt $p_{\text{edit}}$ is activated according to the associated editing state. Specifically, $p_{\text{edit}}$ is included in $p_t$ only when $x_t$ is labeled as \textit{during}. To maintain temporal continuity when editing prompts change across chunks, each chunk $x_t$ attends not only to its current prompt $p_t$, but also to the prompts of the two preceding chunks, $p_{t-1}$ and $p_{t-2}$. This short-term textual context helps reduce abrupt visual transitions caused by prompt switching.

Following the chunk-by-chunk autoregressive generation paradigm, a chunk-causal self-attention pattern is adopted to preserve temporal causality. To enable causal generation without sacrificing training parallelism, the clean latent chunks are concatenated with their noisy counterparts. Let $x_t^{\text{clean}}$ and $x_t^{\text{noisy}}$ denote the clean and noisy latent chunks at time step $t$, respectively. The causal self-attention pattern is defined as
\begin{equation}
\mathcal{S}\left(x_t^{\text{clean}}\right)
=
\left\{
x_j^{\text{clean}} : j \leq t
\right\},
\qquad
\mathcal{S}\left(x_t^{\text{noisy}}\right)
=
\left\{
x_j^{\text{clean}} : j < t
\right\}
\cup
\left\{
x_t^{\text{noisy}}
\right\},
\end{equation}
where $\mathcal{S}(\cdot)$ denotes the set of latent blocks accessible to the corresponding attention query.

Reference images are encoded by the VAE and concatenated with the video latents along the temporal dimension. To incorporate reference images while preserving causality, we introduce a unidirectional gated self-attention mechanism between reference image tokens and video latent tokens. Tokens from each reference image are restricted to attending only to tokens within the same reference image, preventing their representations from being influenced by video tokens or other references. Conversely, the visibility of reference image tokens to a video chunk $x_t$ is gated by its textual context, including $p_t$, $p_{t-1}$, and $p_{t-2}$. Specifically, $x_t$ is allowed to attend to a reference image only when its visible prompts contain semantics referring to that reference. This design explicitly aligns reference conditioning with the textual context of each video chunk, enabling flexible incorporation of reference content at the appropriate generation stage. We further assign reference image tokens a negative temporal RoPE margin $m$ to mitigate direct copy-and-paste behavior. For the $i$-th reference image, all its tokens are assigned a negative temporal RoPE coordinate $m(i+1)$. This negative temporal offset separates reference tokens from the video tokens, encouraging the model to treat reference images as conditioning signals rather than directly copying their spatial content.

\noindent\textbf{Sparse Context.}
To prevent the visual context from growing unboundedly during autoregressive inference, we introduce a Sparse Context mechanism together with sparse attention training. When generating the target chunk $x_t$, the sink chunk $x_0$ and the two most recent chunks, $x_{t-1}$ and $x_{t-2}$, are always retained. The sink chunk preserves the initial world state, while the recent chunks provide short-term temporal context for maintaining generation continuity. In addition, we retrieve the $k$ most relevant chunks from the remaining history and include their KV caches in the active context. As a result, the context budget remains fixed throughout autoregressive generation.

To efficiently retrieve relevant historical chunks, we construct a compact query for the current chunk and compact keys for historical chunks by average pooling their pre-RoPE query and key representations. Given the compact query $\bar{Q}_t$ of the current chunk and the compact key $\bar{K}_j$ of a historical chunk, their relevance is measured using cosine similarity:
\begin{equation}
s_{t,j}
=
\frac{
\langle \bar{Q}_t, \bar{K}_j \rangle
}{
\|\bar{Q}_t\|\,\|\bar{K}_j\|
}.
\end{equation}
The $k$ historical chunks with the highest similarity scores are selected, and the active sparse context is constructed as
\begin{equation}
\mathcal{A}_t
=
\mathcal{S}_t
\cup
\operatorname{TopK}_{j \in \mathcal{H}_t}(s_{t,j}, k)
\cup
\mathcal{R}_t,
\end{equation}
where $\mathcal{S}_t$, $\mathcal{R}_t$, and $\mathcal{H}_t$ denote the sink chunk, recent chunks, and the remaining historical chunks, respectively. During training, a corresponding sparse context strategy is adopted: the sink and recent chunks are always retained, while a random number of additional chunks are sampled from the remaining history. This exposes the model to diverse sparse historical contexts and improves its robustness to the fixed-budget context used during inference.

\subsubsection{Training Curriculum}
We adopt teacher-forcing \citep{williams1989learning} to adapt the bidirectional model for autoregressive video generation. We empirically observe that optimizing only a causal generation objective tends to bias the model toward video continuation, while weakening its responsiveness to diverse input conditions. To mitigate this issue, bidirectional and autoregressive objectives are jointly optimized, allowing the model to retain strong condition-following capability while acquiring causal generation behavior. In addition, an annealed self-resampling strategy \citep{guo2025end} is adopted to improve robustness to error accumulation by progressively exposing the model to its own generated history during training. Finally, we distill the autoregressive model into a few-step model using pCM \citep{wang2024phased}, and further incorporate Self Gradient Forcing \citep{zhuang2026self} to improve the generation quality of the few-step autoregressive model.

\noindent\textbf{Joint Autoregressive and Bidirectional Objectives.}
Optimizing only the causal video generation objective results in relatively weak responsiveness to user inputs. In contrast, the bidirectional model learns to follow editing instructions and reference images within substantially fewer training steps. Following LingBot-World 2.0 \citep{gao2026infinite}, we therefore adopt joint autoregressive and bidirectional training. The two branches share the same model parameters and training data, differing only in their attention patterns. For each video chunk $x_i$, a flow timestep $\tau \sim \mathcal{U}(0,1)$ and Gaussian noise $\epsilon_i \sim \mathcal{N}(0,I)$ are sampled to construct the noisy latent
\begin{equation}
x_i^\tau = (1-\tau)x_i + \tau \epsilon_i.
\end{equation}
Under the autoregressive attention pattern, the flow velocity of each chunk is predicted using only its causal video context and the conditions available up to the current step:
\begin{equation}
\mathcal{L}_{\mathrm{AR}}
=
\mathbb{E}_{x,i,\tau,\epsilon}
\left[
\left\|
v_{\theta}
\left(
x_i^\tau, \tau
\mid
x_{<i}, c_{\leq i}
\right)
-
(\epsilon_i - x_i)
\right\|_2^2
\right].
\end{equation}
Under the bidirectional attention pattern, all video chunks are jointly denoised with full temporal attention:
\begin{equation}
\mathcal{L}_{\mathrm{BI}}
=
\mathbb{E}_{x,\tau,\epsilon}
\left[
\left\|
v_{\theta}
\left(
x^\tau, \tau
\mid
c
\right)
-
(\epsilon - x)
\right\|_2^2
\right].
\end{equation}
The final training objective combines the two branches:
\begin{equation}
\mathcal{L}
=
\mathcal{L}_{\mathrm{AR}}
+
\lambda_{\mathrm{BI}}
\mathcal{L}_{\mathrm{BI}},
\end{equation}
where $\lambda_{\mathrm{BI}}$ controls the relative contribution of the bidirectional objective.

\noindent\textbf{Annealed Self-Resampling.}
At the early stage of training, we adopt standard teacher forcing, where the historical video context consists entirely of clean ground-truth chunks. As training progresses, self-resampling \citep{guo2025end} is introduced by replacing an increasing proportion of ground-truth history with model-generated chunks. An annealing schedule is used to progressively increase the probability of conditioning on self-generated history. By exposing the model to imperfect contexts produced by its own autoregressive rollouts, the model becomes more robust to error accumulation and better maintains stable, high-quality generation over long horizons. The gradual transition from clean ground-truth context to self-generated context also helps stabilize training throughout the curriculum.

\noindent\textbf{Few-Step Distillation.}
Our real-time autoregressive model is trained using a two-stage distillation strategy, consisting of trajectory distillation followed by distribution distillation. In the first stage, we adopt the Phased Consistency Model (PCM) \citep{wang2024phased} as a warm-up to establish few-step generation capability. PCM partitions the teacher denoising trajectory into multiple phases and enforces consistency within each phase, allowing the student to approximate the full denoising process with only a few updates. We use four function evaluations (NFE) for each video chunk. Starting from the PCM-initialized model, we further perform distribution distillation with Self Gradient Forcing (SGF) \citep{zhuang2026self} under the same 4-NFE sampling budget. SGF refines the student's generation distribution using autoregressive rollouts conditioned on its own generated history. Specifically, rollout states are first collected without gradient tracking, followed by a parallel reconstruction pass in which the generated historical latents are treated as detached inputs while their key--value representations are recomputed with gradients enabled. This allows losses from subsequent chunks to supervise both their denoising predictions and the encoding of historical context into causal memory, without backpropagating through the entire sequential rollout. Together, the two stages first establish efficient few-step denoising and then improve generation quality and temporal consistency under self-generated contexts, enabling stable long-horizon autoregressive generation.

%% file: section/4_experiments.tex
\section{Experiments}
\subsection{Comparison on WBench-Editing}
Existing world model benchmarks do not comprehensively evaluate editing capabilities and instead focus primarily on navigation performance and video generation quality. To address this gap, we develop WBench-Editing based on the evaluation framework of WBench \citep{ying2026wbench}. Specifically, approximately 150 cases are redesigned, each spanning 15--30 seconds (240--480 frames), with a subset additionally incorporating reference images. Each case contains one to three editing instructions, and some editing instructions are causally dependent on preceding ones. To evaluate world editing more systematically, we introduce dedicated metrics while retaining most WBench metrics related to navigation and video fidelity. The overall evaluation is organized into six categories: \textit{Editing, Navigation, Quality, Setting, Consistency, and Physical}. Editing capability is assessed from three perspectives: (1) whether the intended modification is correctly executed, (2) the quality of the world after editing, and (3) whether content unrelated to the edit is properly preserved. These editing-related metrics are evaluated through VLM-based question answering.

\begin{figure}[t]
\centering
\includegraphics[width=0.95\linewidth]{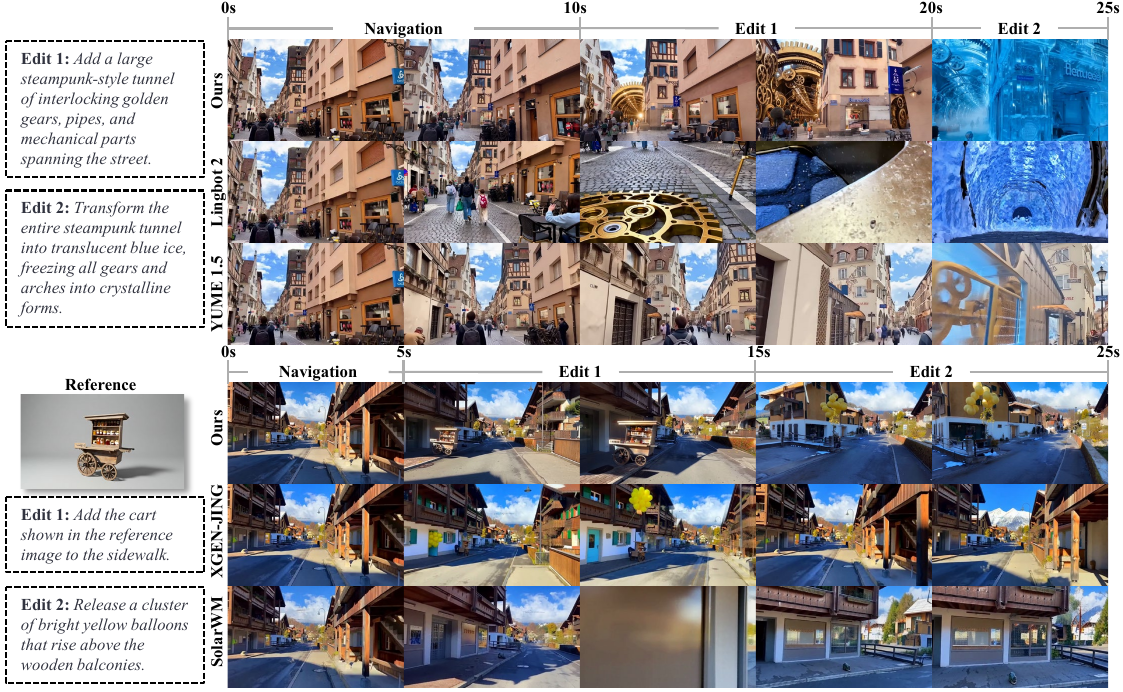}
\caption{Comparison results with video world models on WBench-Editing. The upper panel presents multi-turn world editing with text-only instructions, while the lower panel shows multi-turn editing with reference-image conditioning. Our model follows the editing instructions more accurately while better preserving the consistency of the overall world. Zoom in for the best view.}
\label{fig:comparison}
\end{figure}

Since most existing video world models do not support reference images as interactive conditioning inputs, all models are evaluated using text-only streaming editing instructions for a fair comparison. Table~\ref{tab:wbenchediting} reports the performance of our model and existing world models on WBench-Editing. Because the benchmark requires multiple edits during autoregressive generation, all evaluated methods use their autoregressive variants. Our model achieves the best overall performance with an \textit{Overall} score of 73.8, outperforming the second-best model, YUME 1.5 \citep{mao2026yume1}, by 6.8 points. The advantage is particularly pronounced on the core \textit{Editing} metric, where our model reaches 80.0, exceeding the second-best score of 54.8 by 25.2 points. This substantial margin demonstrates a stronger ability to accurately execute streaming editing instructions while preserving the surrounding world state. Beyond editing accuracy, our model also achieves the best \textit{Physical} score of 67.9, indicating that the edited worlds maintain physical plausibility after content modification. Meanwhile, the model maintains solid performance across \textit{Navigation}, \textit{Quality}, and \textit{Setting}, with scores of 74.3, 74.4, and 60.9, respectively, together with a \textit{Consistency} score of 83.8. These results show that the substantial improvement in editing capability is achieved while preserving the broader world-modeling capabilities required for coherent long-horizon generation. The upper part of Figure \ref{fig:comparison} presents multi-turn editing results from our model and competing methods. Our model responds more accurately to instructions that modify the generated world and better preserves world consistency across successive edits. In contrast, other models tend to interpret editing requests as text-driven events; when new instructions are introduced, they often fail to preserve the existing world state and instead generate substantially different scene content.

\newcommand{\rankone}[1]{\cellcolor{blue!40}#1}
\newcommand{\ranktwo}[1]{\cellcolor{blue!20}#1}
\newcommand{\rankthree}[1]{\cellcolor{blue!10}#1}
\newcommand{\rankfour}[1]{\cellcolor{blue!5}#1}

\begin{table}[t]
\caption{Comparison results on WBench-Editing. All evaluated methods use their autoregressive model variants, and all test cases are evaluated using text-only multi-turn interactive editing instructions. The top four results in each column are highlighted with progressively darker colors.}
\label{tab:wbenchediting}
\centering
\resizebox{1.0\linewidth}{!}
{
\begin{tabular}{l|c|cccccc}
\toprule
Model
& Overall
& Editing
& Navigation
& Quality
& Setting
& Consistency
& Physical \\
\midrule

MatrixGame3 \citep{wang2026matrix}
& 54.0
& 18.1
& 65.7
& 74.0
& 42.8
& 82.5
& 49.7 \\

JoyAI-Echo-1.5 \citep{zhang2026echowm}
& 53.8
& 21.4
& 48.0
& 71.9
& 51.4
& 80.0
& 62.0 \\

minWM(Wan 2.1) \citep{zhao2026minwm}
& 53.2
& 36.2
& 48.1
& 62.8
& 48.6
& 66.7
& 61.5 \\

Astra \citep{zhu2026astra}
& 56.1
& 27.1
& 65.5
& 68.5
& 44.3
& 87.4
& 53.4 \\

Alaya-EVOKE \citep{yin2026alaya}
& 57.7
& 47.2
& 63.6
& 71.2
& 40.2
& 59.8
& 56.6 \\

HY-World 1.5 \citep{sun2025worldplay}
& 56.7
& 21.8
& 62.0
& 64.8
& 57.2
& \rankthree{89.0}
& 61.7 \\

HY-GameCraft \citep{li2025hunyuan}
& 58.7
& 20.2
& 73.8
& \rankthree{75.1}
& 48.4
& \rankfour{88.5}
& 55.7 \\

SANA-WM \citep{zhu2026sana}
& 59.2
& 18.5
& \rankthree{77.7}
& 72.0
& 55.0
& 84.4
& 58.0 \\

ABot-World \citep{jiang2026abot}
& 59.3
& 28.8
& 70.6
& 72.2
& 44.1
& 80.0
& 63.0 \\

Zing-0.5 \citep{zing}
& 63.5
& 33.8
& 64.5
& 73.0
& \rankone{67.7}
& \ranktwo{89.6}
& \rankthree{67.5} \\

AlayaWorld \citep{team2026alayaworld}
& 64.7
& 31.6
& \ranktwo{82.8}
& 63.8
& \rankthree{63.2}
& \rankone{93.6}
& \rankfour{66.7} \\

LingBot-World \citep{team2026advancing}
& \rankfour{66.0}
& \rankthree{54.8}
& 71.3
& \ranktwo{75.8}
& 56.3
& 73.4
& 63.3 \\

DreamX-World \citep{team2026dreamx}
& 65.9
& \rankfour{53.9}
& 70.4
& 73.3
& 55.7
& 82.0
& 63.0 \\

SolarWM(H3) \citep{huang2026solarwm}
& 65.4
& 20.2
& \rankone{86.7}
& \rankfour{75.0}
& \ranktwo{67.1}
& 88.3
& \ranktwo{67.7} \\

LingBot-World 2.0 \citep{gao2026infinite}
& \rankthree{66.1}
& 47.2
& 70.7
& 74.4
& \rankfour{61.3}
& 82.2
& 66.2 \\

YUME 1.5 \citep{mao2026yume1}
& \ranktwo{67.0}
& \ranktwo{54.8}
& 72.9
& \rankone{76.1}
& 53.5
& 84.3
& 62.0 \\

\textbf{EditWorld (Ours)}
& \rankone{73.8}
& \rankone{80.0}
& \rankfour{74.3}
& 74.4
& 60.9
& 83.8
& \rankone{67.9} \\
\bottomrule
\end{tabular}
}
\end{table}

\subsection{Comparison with Reference-Conditioned Video World Models}
Several existing models \cite{jiang2026abot,xgen_jing,huang2026solarwm} support reference images as initial conditioning inputs. We therefore select the WBench-Editing cases that include reference images and compare our model against these methods on this subset. As shown in Table~\ref{tab:wbencheditingref}, our model achieves the best \textit{Overall} score of 74.0 and the highest \textit{Editing} score of 74.6, demonstrating strong reference-conditioned editing capability. More importantly, our model supports flexible reference conditioning throughout autoregressive generation: reference images can be introduced at different interaction stages to modify an already generated world, rather than being restricted to a fixed initial condition. As illustrated in the lower part of Figure \ref{fig:comparison}, XGEN-JING can incorporate the referenced \textit{cart}, but because the reference is provided only at initialization, the timing of the reference-guided modification cannot be controlled precisely. As a result, the \textit{cart} and the yellow balloons appear together instead of following the intended multi-turn sequence in which the cart is introduced first and the balloons are added only in the subsequent edit. SolarWM also supports reference-image conditioning, but exhibits weaker responsiveness to such temporally controlled reference-based edits.

\begin{table}[t]
\caption{Comparison results on WBench-Editing (Reference only).}
\label{tab:wbencheditingref}
\centering
\resizebox{1.0\linewidth}{!}
{
\begin{tabular}{l|c|cccccc}
\toprule
Model
& Overall
& Editing
& Navigation
& Quality
& Setting
& Consistency
& Physical \\
\midrule

ABot-World \citep{jiang2026abot}
& 57.3
& 25.0
& 66.9
& 72.0
& 45.4
& 80.5
& 60.9
\\

SolarWM(H3) \citep{huang2026solarwm}
& 64.6
& 19.8
& \rankone{85.9}
& \rankone{74.8}
& \ranktwo{67.6}
& \rankone{87.5}
& \ranktwo{64.9}
\\

XGEN-JING(BI) \citep{xgen_jing}
& \ranktwo{68.7}
& \ranktwo{63.6}
& 70.4
& 72.4
& \rankone{70.0}
& 79.8
& 62.3
\\

\textbf{EditWorld (Ours)}
& \rankone{74.0}
& \rankone{74.6}
& \ranktwo{80.5}
& \ranktwo{73.3}
& 65.4
& \ranktwo{83.1}
& \rankone{67.3}
\\

\bottomrule
\end{tabular}
}
\end{table}
\begin{table}[t]
\caption{Comparison results on WBench. All baseline results are collected from the official WBench Leaderboard, accessed on September 21, 2026. The top four results in each column are highlighted with progressively darker colors.}
\label{tab:wbench}
\centering
\resizebox{1.0\linewidth}{!}
{
\begin{tabular}{l|c|ccccc}
\toprule
Model
& Average
& Quality
& Setting
& Interaction
& Consistency
& Physical \\
\midrule

Astra \citep{zhu2026astra}
& 63.7
& 67.1
& 59.6
& 66.9
& 73.3
& 51.4 \\

HY-GameCraft \citep{li2025hunyuan}
& 68.2
& 73.0
& 66.6
& 66.3
& 72.6
& 62.4 \\

MatrixGame2 \citep{he2025matrix}
& 68.7
& 73.8
& 67.1
& 80.3
& 65.1
& 57.2 \\

Kairos 3.0 \citep{team2606kairos}
& 70.3
& 74.0
& 70.3
& 64.1
& 82.6
& 60.4 \\

MatrixGame3 \citep{wang2026matrix}
& 71.3
& 75.5
& 63.6
& 83.6
& 74.5
& 59.3 \\

Infinite-World \citep{wu2026infinite}
& 72.8
& 77.0
& 69.3
& 75.4
& 80.0
& 62.1 \\

YUME 1.5 \citep{mao2026yume1}
& 73.3
& 77.6
& 72.4
& 71.4
& 80.1
& 65.2 \\

Astronex-World \citep{zhou2026astronex}
& 73.5
& 78.2
& 73.5
& 63.4
& 83.6
& 68.6 \\

Fantasy-World \citep{dai2026fantasyworld}
& 73.8
& 72.4
& 71.3
& 71.9
& 86.4
& 66.8 \\

InSpatio-World \citep{team2026inspatio}
& 73.9
& 71.5
& 71.4
& 73.2
& 88.4
& 65.2 \\

Genie 3 \citep{genie}
& 73.9
& 75.2
& 72.5
& 73.4
& 82.6
& 65.7 \\

ABot-World \citep{jiang2026abot}
& 74.7
& 76.8
& 71.4
& 84.0
& 79.5
& 61.7 \\

DreamX-World (5B AR) \citep{team2026dreamx}
& 75.0
& 77.5
& 80.8
& 78.6
& 74.9
& 63.3 \\

SANA-WM (4-step AR) \citep{zhu2026sana}
& 76.0
& 79.3
& 76.1
& 82.2
& 80.7
& 61.9 \\

AlayaWorld \citep{team2026alayaworld}
& 76.3
& 79.3
& 69.7
& 80.0
& \rankthree{89.5}
& 63.1 \\

Lyra 2.0 (4-step AR) \citep{shen2026lyra}
& 76.4
& 77.1
& 73.2
& \rankfour{85.6}
& 79.3
& 66.7 \\

Happy Oyster \citep{happyoyster}
& 76.8
& 77.3
& 74.2
& 84.9
& 84.3
& 63.5 \\

LingBot-World (fast) \citep{team2026advancing}
& 77.4
& 79.4
& 77.9
& 79.2
& 84.9
& 65.7 \\

\textbf{EditWorld} (AR, SFT)
& 77.7
& 74.8
& 77.7
& 80.3
& 87.8
& 67.7 \\

HY-World 1.5 (AR distilled) \citep{sun2025worldplay}
& 78.1
& 78.1
& 72.2
& \ranktwo{86.8}
& 86.9
& 66.3 \\

LingBot-World (base-camera) \citep{team2026advancing}
& 78.5
& 78.9
& 72.6
& 80.1
& \rankone{89.9}
& 71.2 \\

LingBot-World v2 \citep{gao2026infinite}
& 79.4
& \rankfour{81.8}
& 76.8
& 82.8
& 86.5
& 69.1 \\

\textbf{EditWorld} (AR, 4-step)
& 79.4
& 75.2
& \rankthree{85.9}
& 78.7
& \rankfour{89.2}
& 68.2 \\

Alaya-EVOKE \citep{yin2026alaya}
& 80.8
& \rankone{82.8}
& 83.8
& 78.6
& 86.9
& \rankfour{72.1} \\

HiDream-O1-World \citep{hiDream}
& 80.9
& 81.0
& 82.2
& 80.0
& 88.0
& \rankthree{73.3} \\

Zing-0.5 \citep{zing}
& 81.0
& 80.6
& 77.8
& 84.2
& 88.5
& \ranktwo{73.8} \\

JoyAI-Echo-1.5 (WM, 4-step) \citep{zhang2026echowm}
& 81.0
& 81.1
& 77.5
& \rankone{87.9}
& 88.3
& 70.1 \\

XGEN-Jing (4-step AR) \citep{xgen_jing}
& 81.0
& 81.3
& \rankone{91.6}
& 78.7
& 83.7
& 69.8 \\

\textbf{EditWorld} (BI, SFT)
& \rankfour{81.2}
& 80.3
& \rankfour{84.6}
& 82.9
& 88.8
& 69.5 \\

JoyAI-Echo-1.5 (WM, BI) \citep{zhang2026echowm}
& \rankthree{81.6}
& 81.5
& 79.4
& \rankthree{86.6}
& \ranktwo{89.8}
& 70.6 \\

XGEN-Jing (BI) \citep{xgen_jing}
& \ranktwo{81.9}
& \ranktwo{82.4}
& \ranktwo{90.3}
& 75.6
& 88.1
& \rankthree{73.3} \\

Alaya-EVOKE-Turbo \citep{yin2026alaya}
& \rankone{82.0}
& \rankthree{81.9}
& 82.1
& 83.9
& 88.1
& \rankone{74.0} \\

\bottomrule
\end{tabular}
}
\end{table}
\subsection{Comparison on WBench}
To demonstrate that our model maintains strong general world-modeling capability, we report comprehensive results on WBench \citep{ying2026wbench}. As shown in Table \ref{tab:wbench}, we evaluate our SFT model under both bidirectional and autoregressive inference patterns. EditWorld (BI, SFT) achieves an \textit{Average} score of 81.2, ranking fourth among all evaluated models and improving upon LingBot-World (base-camera) from 78.5 to 81.2. Notably, substantial gains are observed in \textit{Quality}, \textit{Setting}, and \textit{Interaction}, with \textit{Setting} increasing from 72.6 to 84.6. Under causal autoregressive inference, EditWorld (AR, SFT) achieves an \textit{Average} score of 77.7, remaining competitive with strong open-source and commercial world models. Furthermore, the distilled 4-step autoregressive model improves the \textit{Average} score to 79.4, with particularly strong performance in \textit{Setting} and \textit{Consistency}, reaching 85.9 and 89.2, respectively. Compared with the autoregressive SFT model, this corresponds to a 1.7-point improvement in \textit{Average} and an 8.2-point gain in \textit{Setting}. These results show that EditWorld retains strong general world-modeling performance across different inference patterns, while supporting precise editing and flexible referencing capabilities.

\subsection{Ablation Study}
\begin{figure}[t]
\centering
\includegraphics[width=0.95\linewidth]{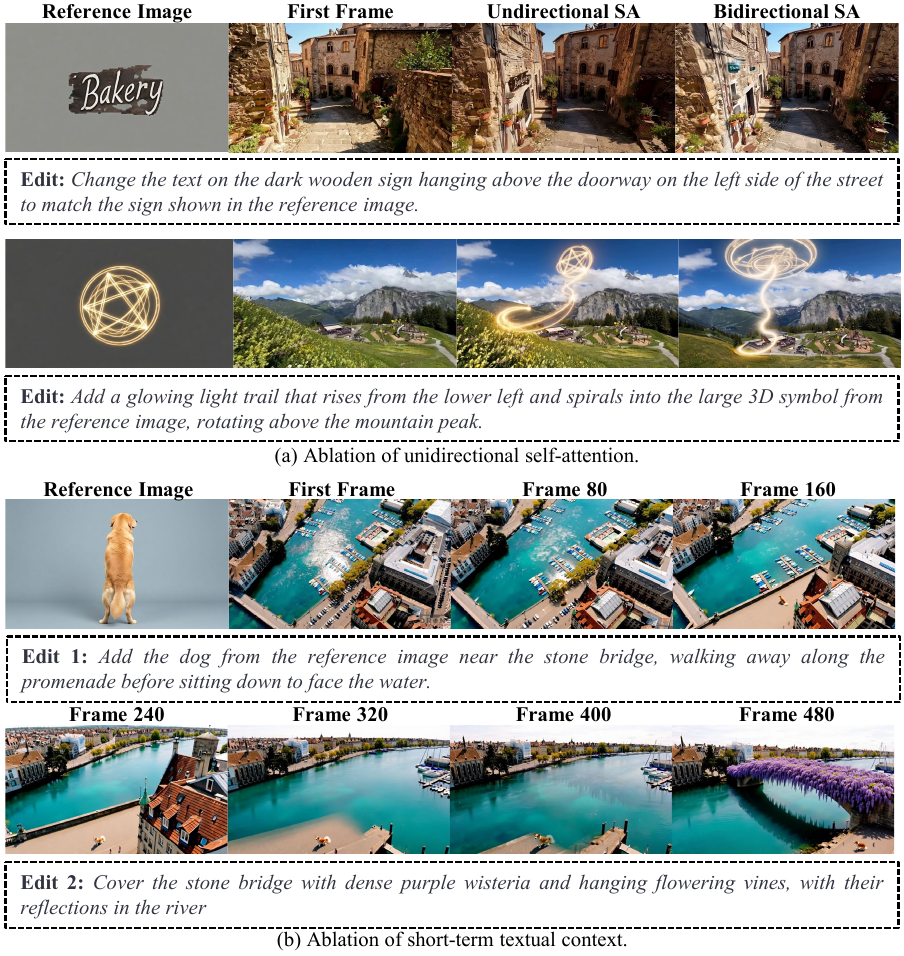}
\caption{(a) Unidirectional reference-video attention better preserves fine-grained reference details than bidirectional patterns. (b) Short-term textual context improves temporal continuity across prompt transitions; without preceding prompts, switching editing instructions causes abrupt scene changes and weaker consistency with the preceding video content.}
\label{fig:ablation}
\end{figure}
\noindent\textbf{Ablation on joint Training Objectives.}
We train our model with joint autoregressive and bidirectional objectives to balance causal video continuation with responsiveness to input conditions. To validate this design, we compare it with an ablated variant trained only with the autoregressive objective. As shown in Table \ref{tab:ablation_objective}, joint AR+BI training improves \textit{Editing Overall} from 69.7 to 80.0 and consistently enhances \textit{Editing Presence}, \textit{Semantic Alignment}, and \textit{Editing Completion}. The largest gain is observed in \textit{Detail Accuracy}, which increases from 34.9 to 53.2, suggesting that the bidirectional objective is particularly important for capturing fine-grained editing requirements. Overall, these results indicate that bidirectional supervision substantially strengthens the model's ability to follow input conditions while preserving autoregressive generation capability.
\begin{table}[h]
\caption{Ablation of the training objective on WBench-Editing. Detailed editing-related metrics are reported for models trained with joint objectives and with the autoregressive objective only.}
\label{tab:ablation_objective}
\centering
\resizebox{1.0\linewidth}{!}
{
\begin{tabular}{l|c|ccccc}
\toprule
Model
& Editing Overall
& Editing Presence
& Semantic Alignment
& Editing Completion
& Detail Accuracy
& Edit Cleanliness \\
\midrule

AR Only
& 69.7
& 82.6
& 79.0
& 73.5
& 34.9
& 78.3 \\

AR + BI
& 80.0
& 93.0
& 89.2
& 86.2
& 53.2
& 78.1 \\

\bottomrule
\end{tabular}
}
\end{table}

\noindent\textbf{Ablation on Unidirectional Self-Attention.}
To prevent video tokens from influencing the representations of reference-image tokens, we adopt unidirectional self-attention between reference and video tokens. To evaluate this design, we construct an ablated variant in which reference and video tokens are mutually visible through bidirectional self-attention. As shown in Figure \ref{fig:ablation}(a), the proposed unidirectional attention better preserves fine-grained details from the reference image. In contrast, bidirectional attention allows video tokens to alter the reference representation, resulting in the loss of reference-specific details and weaker preservation of object identity.

\noindent\textbf{Ablation on Short-term Textual Context.}
To avoid abrupt visual changes when editing prompts switch across chunks, our cross-attention design allows each video chunk to attend to its current prompt as well as the prompts of the two preceding chunks. To evaluate this short-term textual context, we construct an ablated variant in which each chunk attends only to its current prompt. As shown in Figure \ref{fig:ablation}(b), when the editing prompt changes between frames 160 and 240, the ablated model exhibits a pronounced scene shift and an abrupt temporal transition, with the generated environment becoming inconsistent with the preceding video content. In contrast, incorporating prompts from preceding chunks helps preserve the existing world state and enables smoother transitions across editing instructions.

\section{Visualization}
\begin{figure}[t]
\centering
\includegraphics[width=0.9\linewidth]{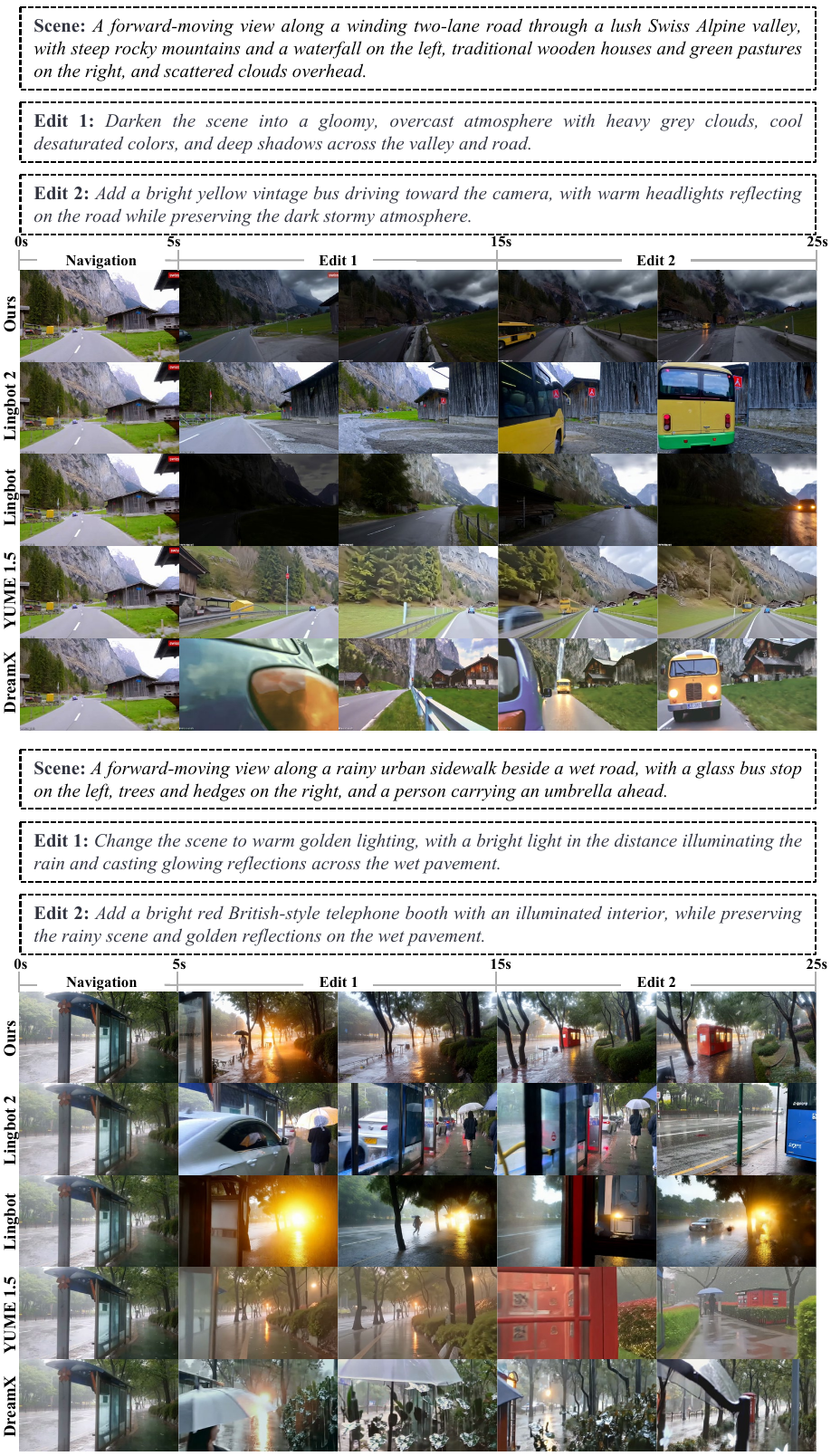}
\caption{Comparison results with existing video world models on WBench-Edit (text only).}
\label{fig:comparison2}
\end{figure}

\begin{figure}[t]
\centering
\includegraphics[width=0.9\linewidth]{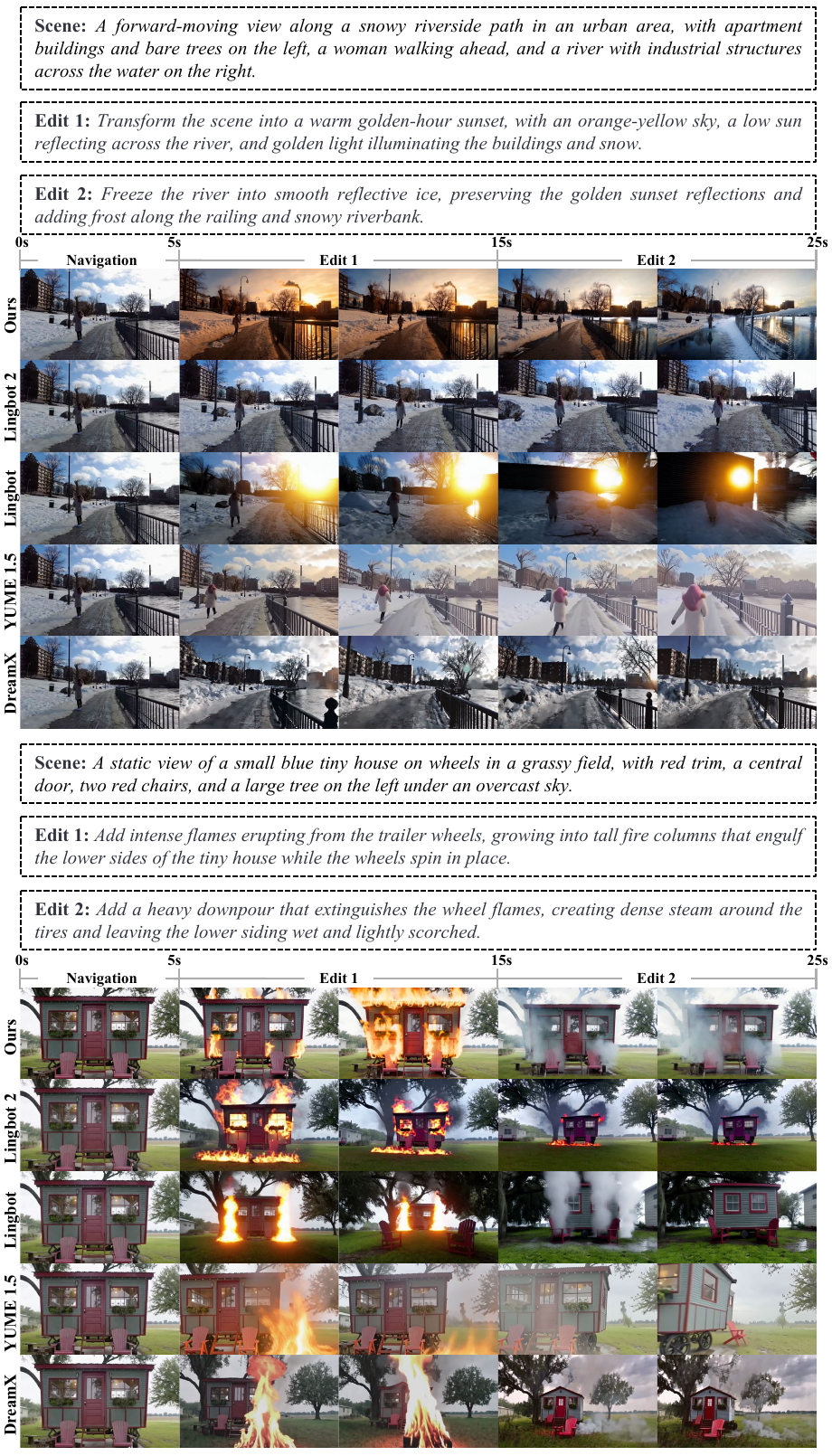}
\caption{Comparison results with existing video world models on WBench-Edit (text only).}
\label{fig:comparison3}
\end{figure}

Figures \ref{fig:comparison2} and \ref{fig:comparison3} present qualitative comparisons between our model and existing video world models on WBench-Editing, including LingBot-World 2.0 \citep{gao2026infinite}, LingBot-World \citep{team2026advancing}, YUME 1.5 \citep{mao2026yume1}, and DreamX-World \citep{team2026dreamx}. As shown, our model follows the specified editing instructions more accurately while preserving smooth camera motion and stronger scene consistency across multiple editing rounds. In contrast, competing models often exhibit incomplete edits, unintended changes to unrelated scene content, or substantial deviations from the preceding world state. These qualitative results further demonstrate the advantage of our model in accurately executing sequential world modifications while maintaining consistency over time.

\begin{figure}[t]
\centering
\includegraphics[width=0.9\linewidth]{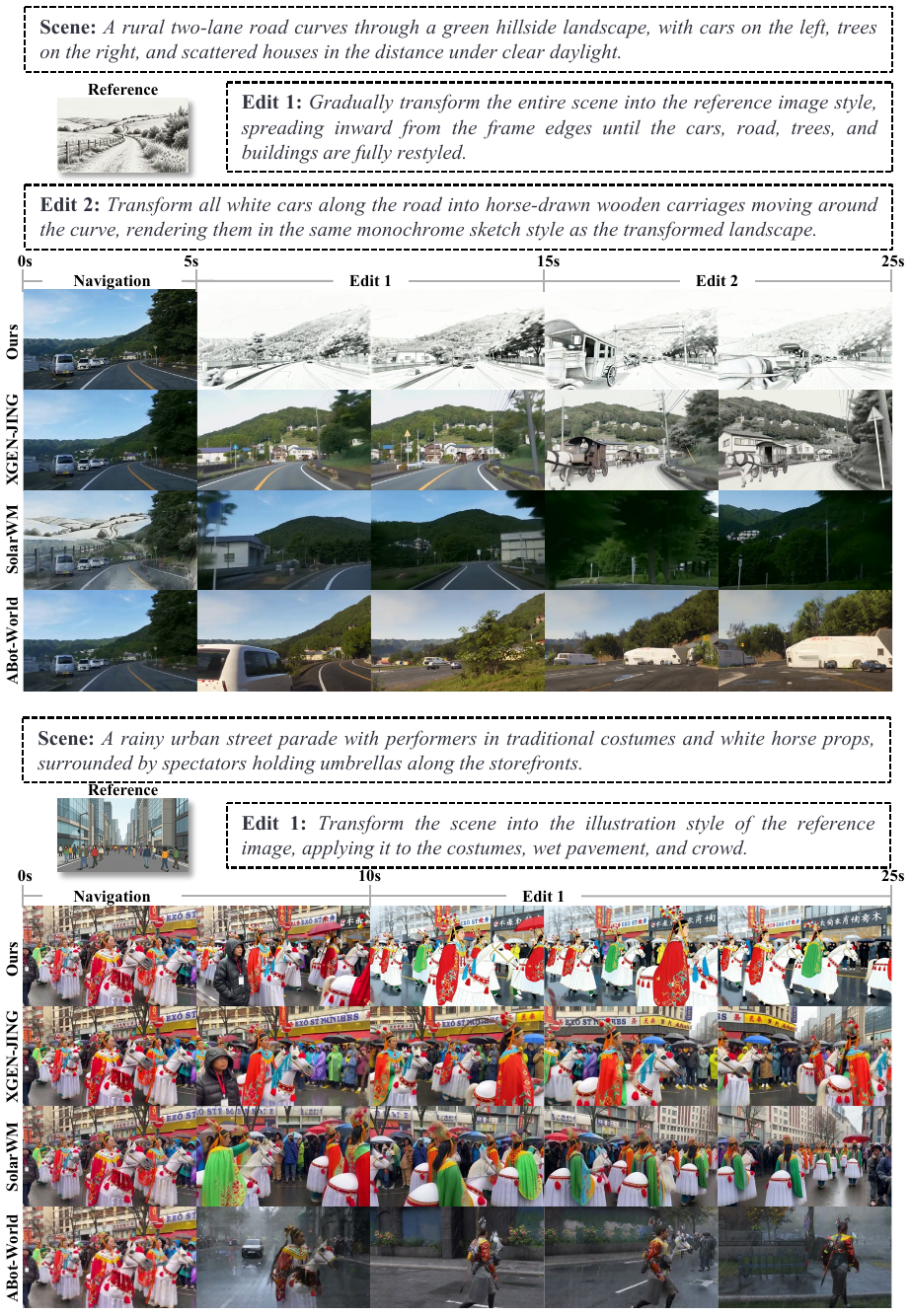}
\caption{Comparison results with video world models on WBench-Edit (with reference image).}
\label{fig:comparisonref1}
\end{figure}

\begin{figure}[t]
\centering
\includegraphics[width=0.9\linewidth]{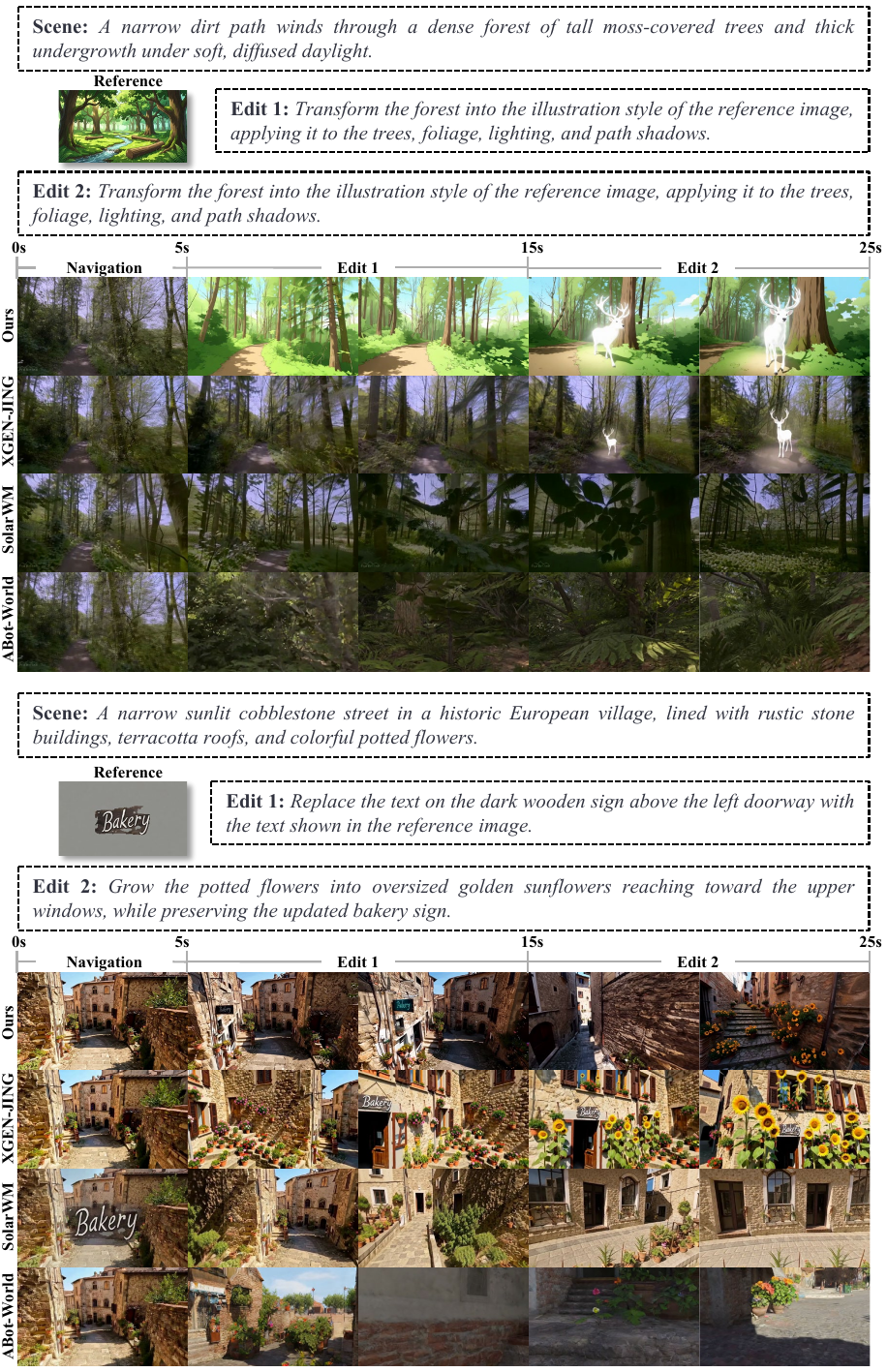}
\caption{Comparison results with video world models on WBench-Edit (with reference image).}
\label{fig:comparisonref2}
\end{figure}

Figures \ref{fig:comparisonref1} and \ref{fig:comparisonref2} present qualitative comparisons on WBench-Editing cases with reference-image conditioning. Our model accurately incorporates content from the provided references, including appearance, style, object identity, and fine-grained visual attributes, while preserving the surrounding world context across successive edits. More importantly, we can introduce reference images flexibly at different interaction turns rather than restricting them to a fixed initial condition. This allows users to inject new reference content into an already generated world and combine reference-guided modifications with subsequent editing instructions. In contrast, existing reference-conditioned models often rely on the reference image only at initialization or show weaker control over when and how the referenced content is introduced. These results highlight the advantage of our model in flexible reference conditioning for multi-turn interaction world editing.

\begin{figure}[t]
\centering
\includegraphics[width=0.95\linewidth]{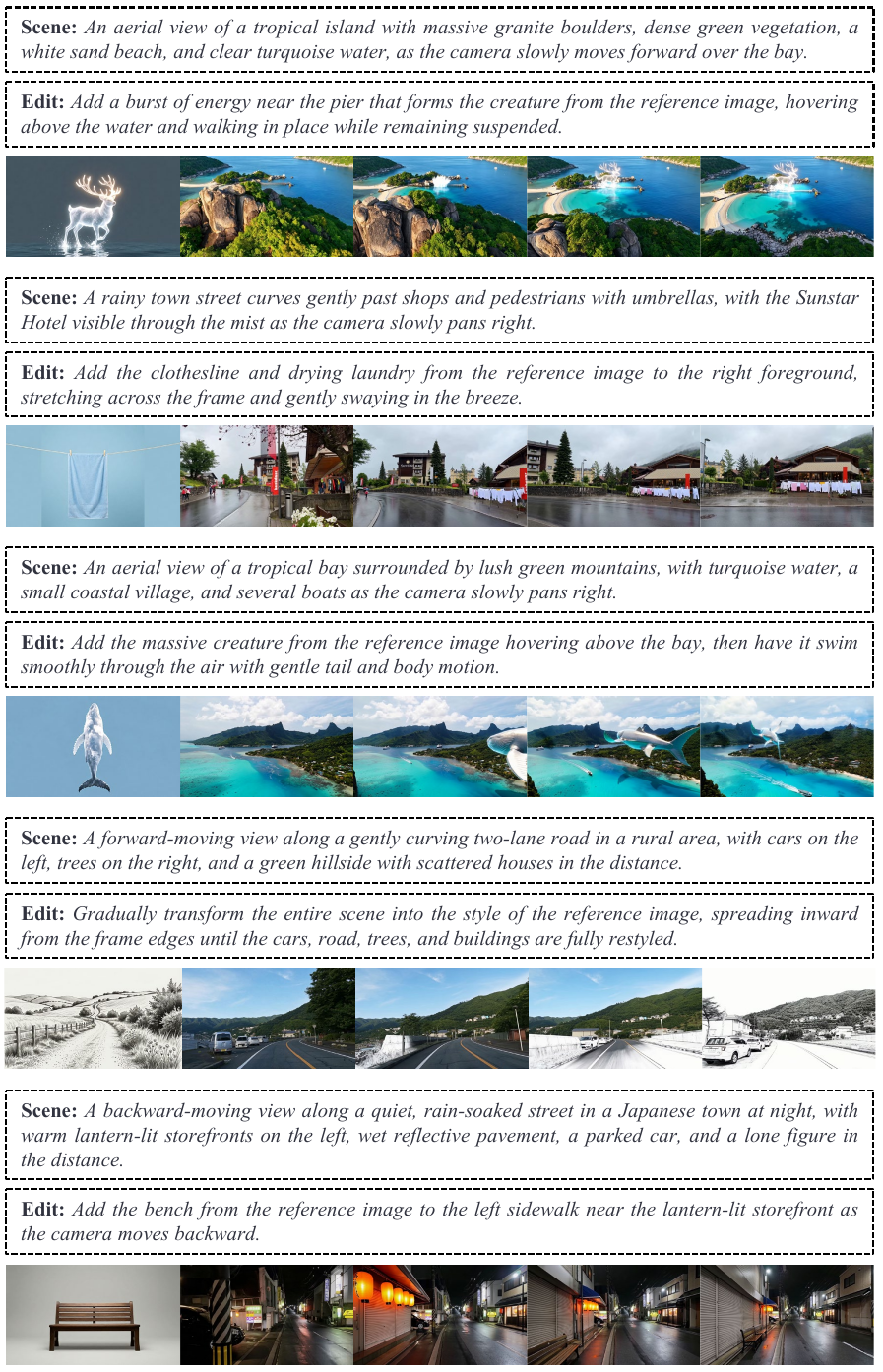}
\caption{Visualization results of our model. The first image on the left is the reference input.}
\label{fig:show1}
\end{figure}

\begin{figure}[t]
\centering
\includegraphics[width=0.95\linewidth]{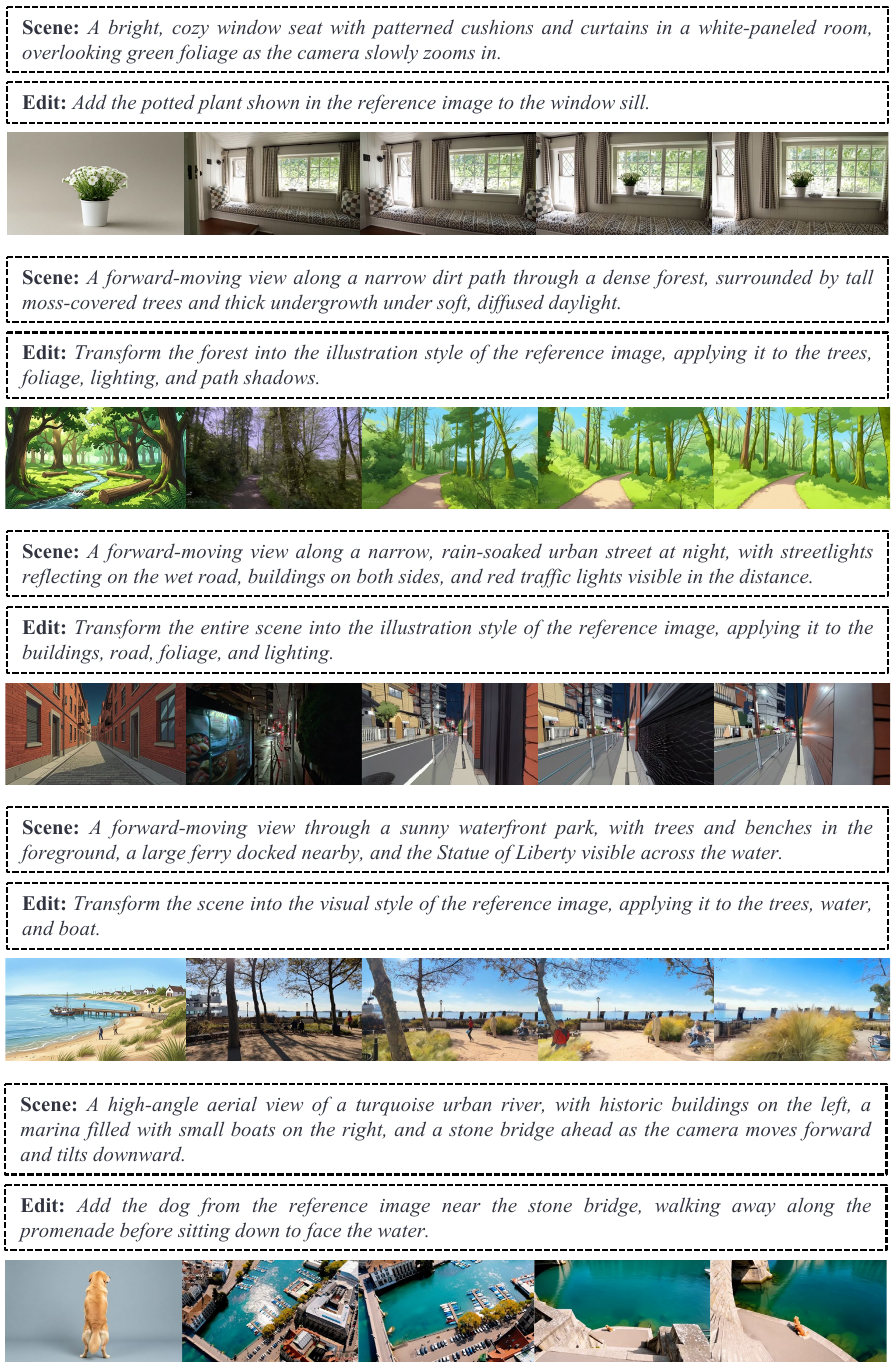}
\caption{Visualization results of our model. The first image on the left is the reference input.}
\label{fig:show2}
\end{figure}

\begin{figure}[t]
\centering
\includegraphics[width=0.95\linewidth]{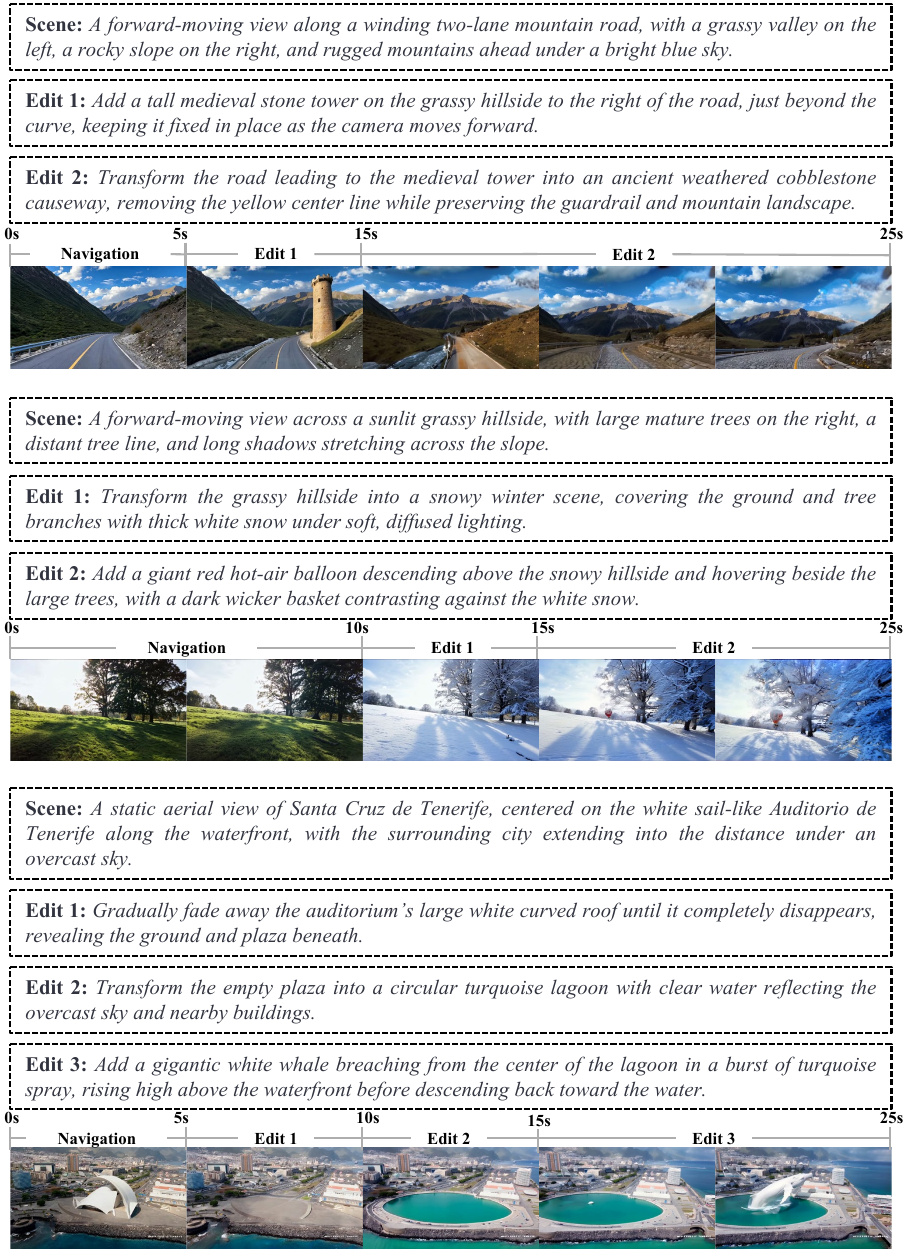}
\caption{Visualization results of multi-turn interactive editing of our model.}
\label{fig:show3}
\end{figure}

\begin{figure}[t]
\centering
\includegraphics[width=0.95\linewidth]{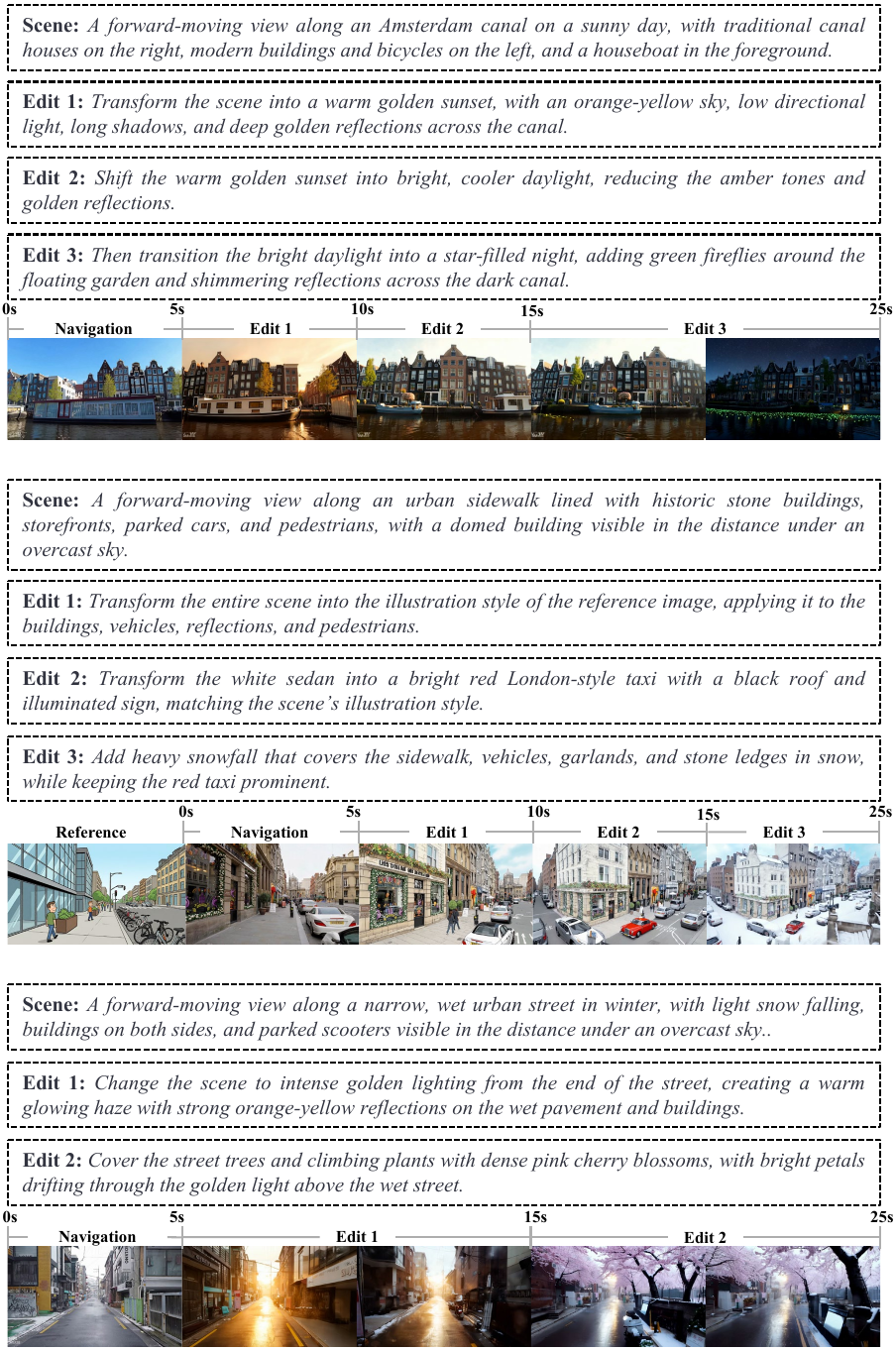}
\caption{Visualization results of multi-turn interactive editing of our model.}
\label{fig:show4}
\end{figure}

Figures \ref{fig:show1} and \ref{fig:show2} present generation results conditioned on reference images, showing that our model can accurately incorporate referenced objects into the generated world and flexibly modify scene appearance according to the provided visual guidance. Beyond object-level conditioning, the model also supports more abstract reference-based control, such as transferring the visual style of a reference image to the generated world. Figures \ref{fig:show3} and \ref{fig:show4} further demonstrate multi-turn interactive editing. Across successive editing rounds, our model accurately follows the requested modifications while preserving scene elements unrelated to the current edit.

%% file: section/5_conclusion.tex
\section{Conclusion}
This work presents EditWorld, a video world model that moves beyond navigation toward precise world editing and flexible reference-based control. EditWorld enables users to modify generated worlds through streaming instructions, actions, and reference images while maintaining stable long-horizon generation. To support this capability, we develop dedicated model components, training strategies, and a data pipeline tailored to world modification, and introduce WBench-Editing for systematic evaluation of streaming editing. Experiments show that EditWorld substantially improves editing performance while preserving strong general world-modeling capability.
We hope this work contributes to video world models that users can continuously and flexibly refine, reshape, and extend through interaction.